\documentclass{article}
\usepackage{spconf,amsmath,graphicx,amssymb,booktabs,multirow,multicol,arydshln,makecell,float}

\title{An adaptive Plug-and-play network for few-shot learning}
\name{Hao Li$^1$, Li Li$^{1\ast}$, Yunmeng Huang$^1$, Ning Li$^1$, Yongtao Zhang$^2$\thanks{$\ast$ Corresponding author.}}
\address{$^1$School of Electronic and Information Engineering, Beihang University\\
$^2$College of Mechanical \& Electrical Engineering, 
Nanjing University of Aeronautics and Astronautics}
\begin{document}
% \ninept
%
\maketitle
\begin{abstract}
Few-shot learning (FSL) requires a model to classify new samples after learning from only a few samples. While remarkable results are achieved in existing methods, the performance of embedding and metrics determines the upper limit of classification accuracy in FSL. The bottleneck is that deep networks and complex metrics tend to induce overfitting in FSL, making it difficult to further improve the performance. Towards this, we propose plug-and-play model-adaptive resizer (MAR) and adaptive similarity metric (ASM) without any other losses. MAR retains high-resolution details to alleviate the overfitting problem caused by data scarcity, and ASM decouples the relationship between different metrics and then fuses them into an advanced one. Extensive experiments show that the proposed method could boost existing methods on two standard dataset and a fine-grained datasets, and achieve state-of-the-art results on \emph{mini}-ImageNet and \emph{tiered}-ImageNet.
\end{abstract}
\section{Introduction}
\label{sec:intro}
The purpose of FSL is to learn a classifier that can recognize unseen categories from a small number of samples. A mainstream mechanism \cite{1597116,lake2011one} is to learn transferable knowledge from seen categories, and then use this transferable knowledge to build a classifier, and finally apply the classifier to unseen categories. Following this mechanism, recently great advances have been made in metric-based methods for FSL, achieving state-of-the-art results.

The pipeline of metric-based methods is to learn advanced deep embedding module and then measure the similarity between query sample and a few support samples \cite{ProtoNet}. 
The seminal work could date back to \cite{SiameseNN}, which propose Siamese Network for obtaining image representations with $L_1$ distance as similarity metric. Then Matching Network \cite{MatchingNetwork} introduce attention mechanism and cosine metric. Prototypical Network \cite{ProtoNet} propose mean vector as the corresponding class prototype representation and adopt Euclidean distance.
On the basis of the above methods, recent approaches can be broadly divided into two categories. The first exploits Gaussian distribution \cite{ADM,FreeLunch}, discrete probability \cite{DeepEMD,DeepBDC-CVPR2022}, Ridge regression feature reconstruction \cite{Reconstruction}, and attention mechanism \cite{CarlDoersch2020CrossTransformersSF,FEAT,HCTransformer} for advanced embedding representations. And the second category aims to obtain progressive similarity metric by learnable metric modules \cite{RelationNet,HongguangZhang2018PowerNS}, covariance matrices \cite{CovaMNet,Localization}, nearest neighbors \cite{DN4}, and multiple-metric fusion \cite{BSNet}. Remarkable results have been achieved in above methods. However, deep networks are prone to overfitting when data scarcity, making it a challenge to continue improving performance of FSL. To solve this challenge, the proposed approach presents a pioneering solution consisting of two ways.
 
\begin{figure}[!t]
\centering
\includegraphics[width=0.85\linewidth]{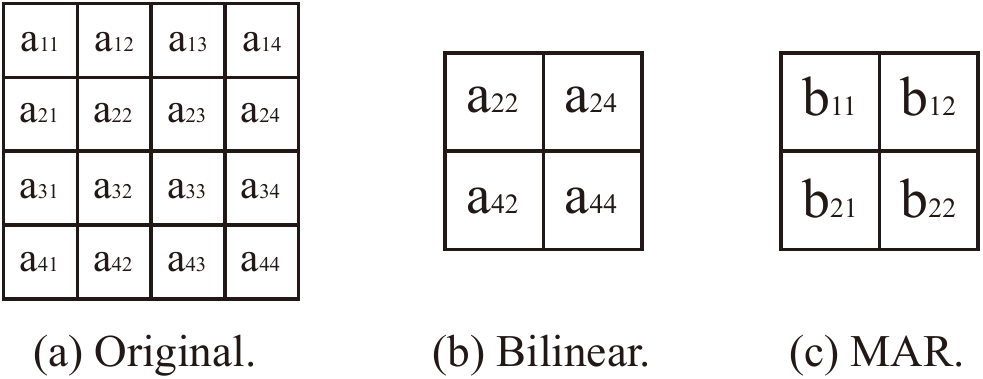}
\caption{Diagram of bilinear and MAR resizing. As shown in (b), bilinear resizing is equivalent to sampling the original image in (a). However, as shown in (c), MAR resizing yields new pixels.}
\label{fig:bilinear_vs_mar}
\end{figure}

The first is adding learnable preprocessing module in front of backbone network. Learnable preprocessing modules have been early shown could improve visual recognition \cite{Deconvolution}. Most approaches explore from super resolution \cite{Namboodiri,Haris}, image decompression \cite{Luo}, image denoising \cite{Diamond}, image dehazing \cite{Li} to improve the performance of specific task of visual recognition. \cite{liu2019transferable} makes great progress, proposing a generic framework effective on multiple tasks. It cannot be used directly for FSL because it requires the introduction of specific loss and do not take into account FSL characteristics. For FSL, due to data scarcity, it is crucial for the framework to obtain enough discriminative details. However, as shown in Fig. \ref{fig:bilinear_vs_mar}, traditional resizing methods directly downsample high-resolution images to a uniform size, thus details with discriminative information in high-resolution images may be lost.
To solve this problem, we propose MAR, a learnable model-adaptive resizer. MAR incorporates the spatial information of surrounding pixels during resizing, so that the details of high-resolution images can be effectively preserved. In addition, MAR does not introduce any other losses thus can be flexibly instantiated in existing methods.

The second is to adaptively fuse existing metrics to obtain advanced discriminative one. Existing methods mainly focus on finding advanced discriminative higher-order metrics, and there has been little research on metric fusion. As the first method to fuse multiple measures into one, \cite{BSNet} proves that multiple simple metrics outperform a few single high-order metrics by manually presetting the weights. However, manual presets are based on sparse priors, which cannot deeply mine the relationship among various metrics. To solve the problem, we propose an adaptive metric ASM. ASM can adapt to different combinations of metrics by learning the weights. Extensive experiments show that the proposed ASM outperforms the method in \cite{BSNet}.

\begin{figure}[!t]
\centering
\includegraphics[width=0.9\linewidth]{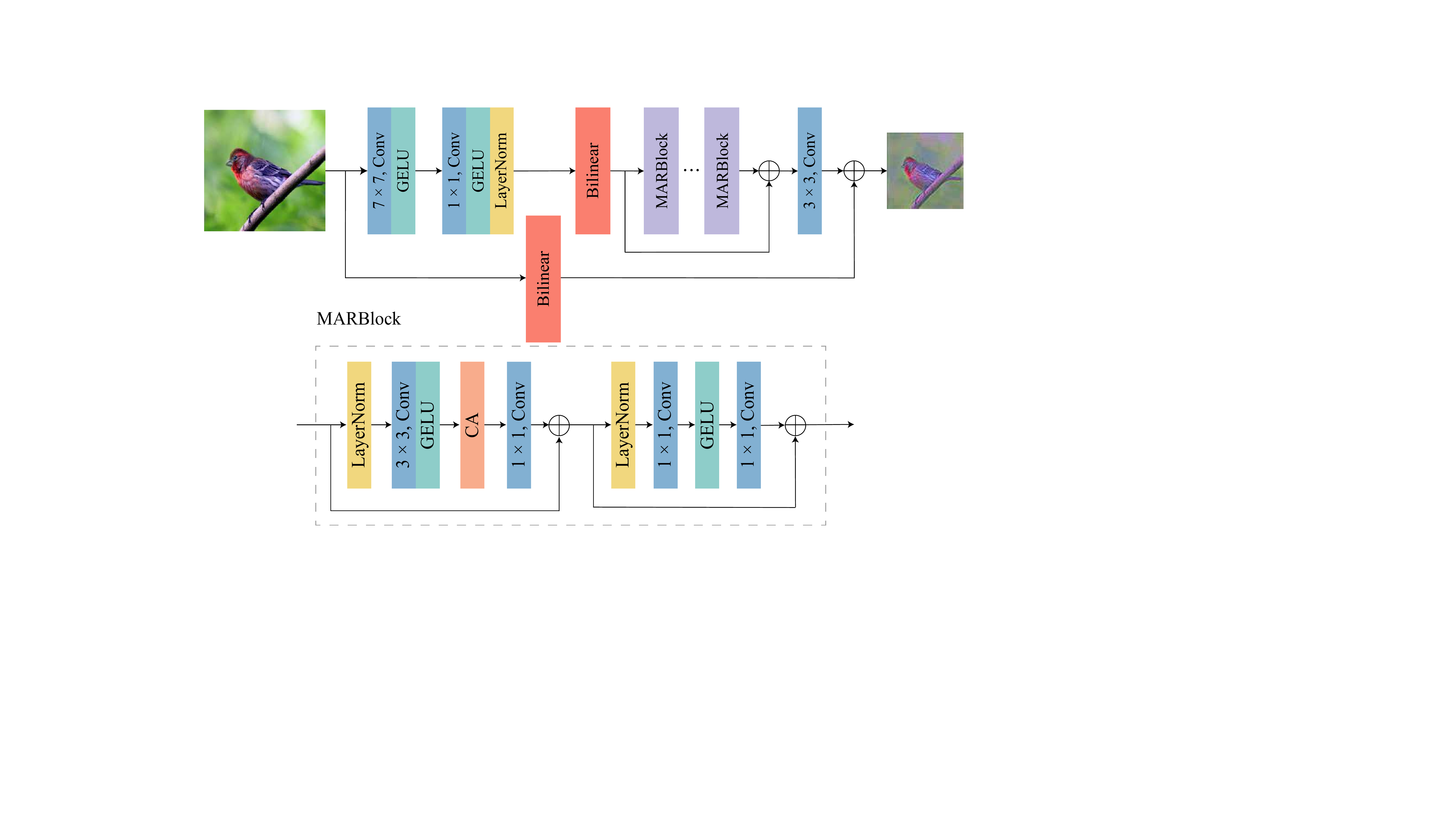}
\caption{The architecture of the proposed MAR. CA denotes channel attention module.}
\label{fig:mar}
\end{figure}

\section{Proposed Method}
\label{sec:format}
In this section, we first outline metric-based FSL, and then introduce the details of proposed framework.
\subsection{FSL setting}
In standard FSL, the dataset $\mathcal{C}$ consists of training set $\mathcal{C}_{train}$, validation set $\mathcal{C}_{val}$ and test set $\mathcal{C}_{test}$. Their classes do not have intersection.
A specific task $\mathcal{T}$ in FSL is denoted as $N$-way $K$-shot classification, which represents a random sampling of $N$ classes from the support set $\mathcal{D}_{support}$ with $K$ samples per class.
Given a query set $\mathcal{D}_{query}$, the porpose is to correctly classify unlabeled samples in it by learning the knowledge in the support set $\mathcal{D}_{support}$.
We denote the support set as $\mathcal{D}_{support} = (\textbf{x}_i, \textbf{y}_i)_{i=1}^{N, K}$, with the instance as $\textbf{x}_i \in \mathbb{R}^\mathcal{D}$ and one-hot vector label $\textbf{y}_{i} = \{0,1\}^N$. The value of $N$ is generally $1$ and $5$.

The first step of metric-based methods is to obtain a classifier by regular classification training on $\mathcal{C}_{train}$, and then take specific tasks from $\mathcal{C}_{train}$ to fine-tune.
The final goal is to obtain a classifier $f_{\theta}(\cdot)$ that performs well on unseen classes $\mathcal{C}_{test}$. Mathematically,
\begin{equation}
\hat{\textbf{y}}_j = f_{\theta}(\textbf{x}_j; \mathcal{C}_{train}), \textbf{x}_j \in \mathcal{C}_{test} 
\label{eq0}
\end{equation}
where ${\textbf{x}}_j$ denotes the unseen sample and $\hat{\textbf{y}}_j$ denotes the predicted label. 

\subsection{Model-adaptive resizer}
The proposed MAR is shown in Fig. \ref{fig:mar}. For large receptive field and less parameters, the convolution kernels of the first two layers are 7 × 7 and 1 × 1. And GELU is used for nonlinear activation. Then features are resized by bilinear to a uniform size. Output features are fed into $n$ MARBlocks, where MARBlock is a module for feature enhancement. After the features plus residual are convolved with a 3 × 3 kernel and then summed with original downsampled residual, the MAR output is obtained. As shown in Fig. \ref{fig:bilinear_vs_mar}, the conventional resizer is only able to extract partial information of the high-resolution image. However, the proposed MAR can extract partial information while preserving some key details in high-resolution images.
\subsection{Adaptive similarity metric}
In general, inputs are all transformed into vectors of the same length after embedding, and prediction is achieved by similarity measurement on the vectors from $\mathcal{D}_{support}$ and $\mathcal{D}_{query}$. And the widely adopted metrics are Euclidean distance and cosine similarity in existing methods. Almost all methods select one of them, and few have explored the fusion of them. As unique to our knowledge, BSNet \cite{BSNet} innovatively explores metric fusion roughly by manually presetting the weights. To further extend the work of BSNet from an adaptive perspective, we propose ASM. Specifically, given a FSL task $\mathcal{T}$, $\mathcal{C}_k$ is used to denote the sample set of class $k$ in the task. The prototype of class $k$ is represented as
\begin{equation}
{\omega}_k = \frac{1}{\mathcal{C}_k} \sum_{(\textbf{x}_i, \textbf{y}_i) \in \mathcal{C}_k} \Psi(\textbf{x}_i)
\label{eq4}    
\end{equation}
where $\Psi(\cdot)$ represents embedding module. For the query sample $\textbf{x}_j$ in task, similarity with $\mathcal{C}_k$ can be denoted as
\begin{equation}
\mathrm{sim}(\Psi(\textbf{x}_j), {\omega}_k) = \alpha \cdot \mathrm{euc}(\Psi(\textbf{x}_j), {\omega}_k) + \beta \cdot \mathrm{cos}(\Psi(\textbf{x}_j), {\omega}_k)
\label{eq5}
\end{equation}
where $\alpha$ and $\beta$ denotes learned weights, $\mathrm{sim}(\cdot)$ denotes ASM, $\mathrm{euc}(\cdot)$ denotes Euclidean distance and $\mathrm{cos}(\cdot)$ denotes cosine similarity, respectively. 
% figure2
\begin{figure*}[t]
\centering
\includegraphics[width=0.85\linewidth,scale=1.00]{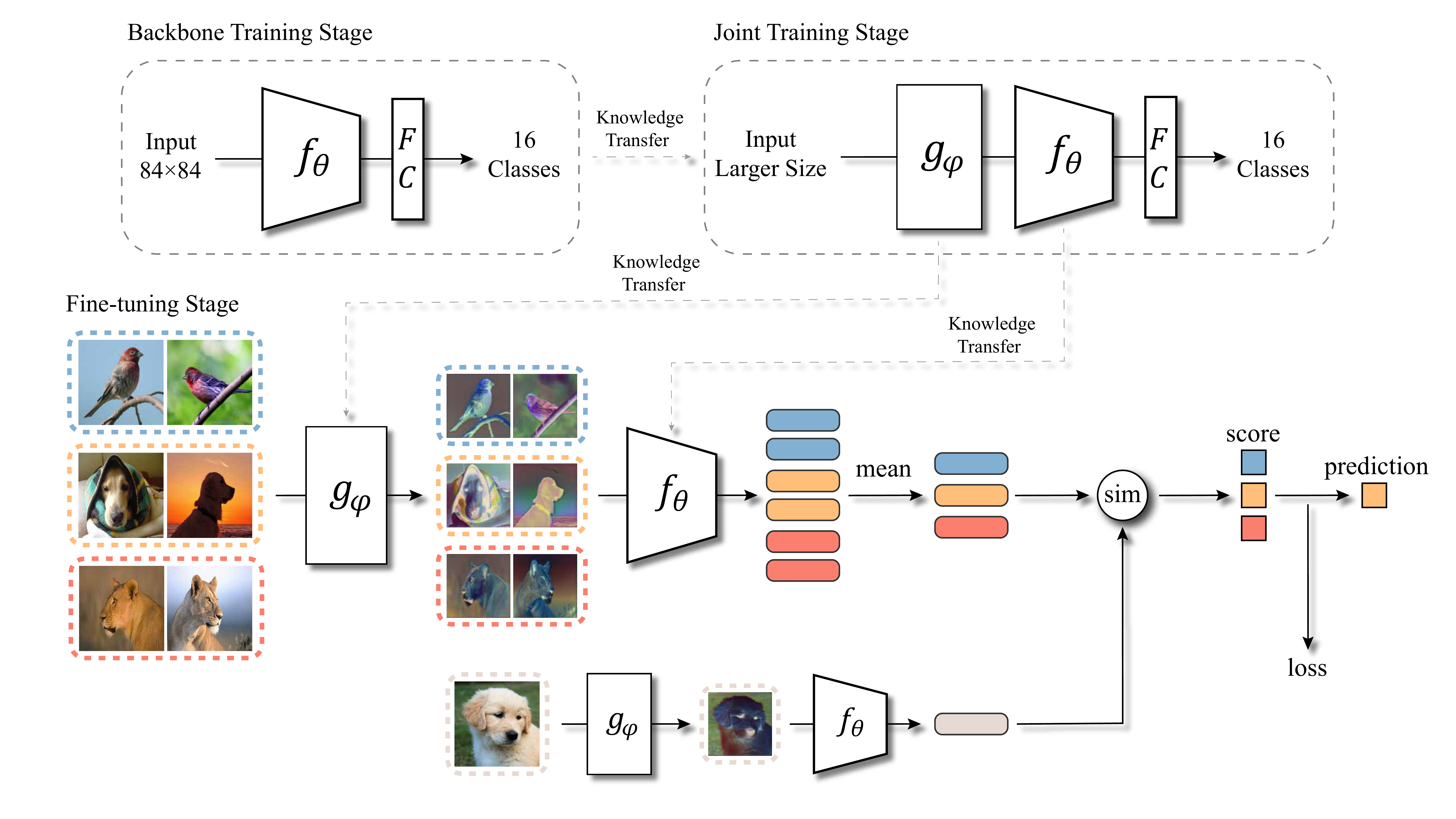}
\caption{Illustration of the proposed framework in 3-way 2-shot setting. $f_\theta(\cdot)$ represents backbone network, $g_\phi(\cdot)$ represents MAR and $sim(\cdot)$ is ASM. Training consists of three stages: backbone training, joint training and fine-tuning.}
\label{fig2}
\end{figure*}
\subsection{Implementation of proposed method}
In our method, as shown in Fig. \ref{fig2}, training is divided into three stages: backbone training, joint training and fine-tuning. In backbone training stage, we only train the backbone network $f(\cdot)$ in traditional classification way to learn the parameters $\theta_{1}$, obtaining the preliminary model $f_{\theta_1}(\cdot)$. Then, in joint training stage, MAR, denoted as $g(\cdot)$ is instantiated in front of $f_{\theta_1}(\cdot)$. This stage learns parameters $\phi_1$ and $\theta_2$ in the same way as first stage. The obtained model are denoted as $g_{\phi_1}(\cdot)$ and $f_{\theta_2}(\cdot)$, respectively. Finally, in fine-tuning stage, we train the model obtained in the second stage in meta manners to obtain the final parameters $\phi$ and $\theta$, i.e., the embedding module $\Psi(\cdot)=f_{\theta} \circ g_{\phi}(\cdot)$.

Specifically, in traditional classification way, training is done by adding a fully-connected (FC) layer after CNN. Then FC layer is removed and we randomly sample several tasks $\mathcal{T}$ from $\mathcal{D}_{query}$ in validation set $\mathcal{C}_{val}$ to save the best model.
In meta manner, i.e. fine-tuning stage, a large number of $(\textbf{x}_i, \textbf{y}_i)$ are taken from $\mathcal{D}_{support}$. Then $\textbf{x}_i$ are fed to resizer $g_{\phi_1}(\cdot)$ to get the resized $\textbf{x}_i$ (denoted as $\widetilde{\textbf{x}}_i$), and $\widetilde{\textbf{x}}_i$ passes backbone $f_{\theta_2}(\cdot)$ to get the embedding for similarity measure and loss calculation. Mathematically, backbone training stage can be written as
\begin{equation}
\mathop{\arg\min}\limits_{\theta_1} \sum_{(\textbf{x}_{j}, \textbf{y}_{j}) \in \mathcal{D}_{query}}l{\ }(\textbf{y}_j, f_{\theta_1}(\textbf{x}_j; \mathcal{D}_{support}))
\label{eq1}
\end{equation}
where $(\textbf{x}_j, \textbf{y}_j)$ denotes the sample pairs taken from validation set $\mathcal{C}_{val}$, and $l(\cdot)$ represents the loss between the prediction and the actual label. 
%And $f_{\theta_1}(\cdot)$ is obtained by minimizing the loss through multiple epochs of training. 
Similarly, joint training stage can be marked as
\begin{equation}
\mathop{\arg\min}\limits_{\theta_2,\phi_1} \sum_{(\textbf{x}_{j}, \textbf{y}_{j}) \in \mathcal{D}_{query}}l{\ }(\textbf{y}_j, f_{\theta_2} \circ g_{\phi_1}(\textbf{x}_j; \mathcal{D}_{support}))
\label{eq2}
\end{equation}
where $\theta_2$, $\phi_1$ are learned parameters, and other terms are consistent with Eq. \ref{eq1}. Fine-tuning stage is the same as the joint training, except that the learned objectives are $\theta$ and $\phi$.

Overall, $\alpha$, $\beta$, $\theta$ and $\phi$ are updated with cross-entropy loss. Finally, predictions are mapped from embedding space to probabilities via softmax. Concretely, the probability of query sample $\textbf{x}_j$ belonging to category $k$ is
\begin{equation}
    p(\hat{\textbf{y}}_j=k|\textbf{x}_j) = \frac{\exp(\mathrm{sim}(\Psi(\textbf{x}_j), {\omega}_k))}{\sum_{\textbf{x}_j \in \mathcal{D}_{query}} \exp(\mathrm{sim}(\Psi(\textbf{x}_j), {\omega}_k))}
\end{equation}
and the loss can be calculated by
\begin{equation}
    \mathcal{L}=-\frac{1}{ \left|\mathcal{D}_{query}\right|}\sum_{(\textbf{x}_j,\textbf{y}_j) \in \mathcal{D}_{query}} \textbf{y}_j \log(p(\hat{\textbf{y}}_j=k|\textbf{x}_j))
\end{equation}
%After many steps, $\phi_1$ and $\theta_2$ are updated and finally we get $\phi$ and $\theta$. 
% Formally, the pipeline can be written as
% \begin{equation}
% \hat{\textbf{y}}_j = f_{\theta} \circ g_{\phi}(\textbf{x}_j; \mathcal{D}_{support}), \textbf{x}_j \in \mathcal{D}_{query} \label{eq1}
% \end{equation}
% where $f_{\theta}(\cdot)$ and $g_{\phi}(\cdot)$ are obtained by training on $\mathcal{D}_{support}$, and then $\textbf{x}_j \in \mathcal{D}_{query}$ is fed into $g_{\phi}(\cdot)$ and $f_{\theta}(\cdot)$ to get the prediction of label of $\textbf{x}_j$.

\section{EXPERIMENTS}
\label{sec:pagestyle}
\subsection{Datasets and experiment settings}
Our experiments are conducted on datasets \emph{mini}-ImageNet \cite{MatchingNetwork}, \emph{tiered}-ImageNet \cite{tieredimagenet} and CUB \cite{wah2011caltech}. For \emph{mini}-ImageNet and \emph{tiered}-ImageNet we adopt ResNet-12 as the backbone network, and for CUB we employ Conv-4 and ResNet-12 (Conv-4 is the same setting as in \cite{BSNet}). In training of ResNet-12, we utilize Adam optimizer with weight decay 0.0005, and learning rate starts from 0.0001 and drops by a factor of 0.1. For Conv-4, we use SGD optimizer with a momentum of 0.9 with learning rate starting at 0.0002 and decreasing by a factor of 0.5.

%sota
\begin{table}[!t]\small   
\begin{center}
\begin{tabular}{lccccc}   
\toprule
\multirow{2}{*}{Model} & 
\multicolumn{2}{c}{\emph{mini}-ImageNet} & \multicolumn{2}{c}{\emph{tiered}-ImageNet} \\
& 1-shot & 5-shot  & 1-shot & 5-shot\\
\midrule
    
    DeepEMD \cite{DeepEMD} & 65.91& 82.41& 71.16 &  86.03  \\ 
    
    BML \cite{BML}& 67.04 & 83.63 & 68.99 & 85.49 \\
    
    Baseline++ \cite{chen2018a}$^\dag$ &  63.15& 81.77  & 67.97 & 84.43 \\
    ProtoNet \cite{ProtoNet}$^\dag$ & 62.26 & 80.19 & 67.47 & 82.07 \\
    
    FEAT \cite{FEAT}$^\dag$ &  66.67 & 82.01 & 70.97  & 84.87  \\
    
    Meta-Baseline \cite{YinboChen2020MetaBaselineES}$^\dag$ & 64.91 & 80.78  & 65.77 & 83.21 \\
    
    Meta DeepBDC \cite{DeepBDC-CVPR2022}$^\dag$ & 67.37 & 84.39 & 72.68 & 87.87 \\
\midrule
    Ours &  65.19 & 82.38 & 69.01 & 85.17 \\
    
    FEAT + Ours &  67.41 & 82.98 & 71.74 & 85.81 \\
    
    Meta-Baseline + Ours & 65.11 & 81.67 & 68.81 & 85.34\\
    
    Meta DeepBDC + Ours &  \textbf{68.24} & \textbf{84.45} & \textbf{73.71} & \textbf{87.91} \\
\bottomrule   
\end{tabular}   
\caption{Results on \emph{mini}-ImageNet and \emph{tiered}-ImageNet at 5-way setting. The best results are marked in \textbf{bold black}.}
\label{table:sota}

%2
\begin{tabular}{ccccccc}   
\toprule
 \multirow{2}{*}{Model} & \multirow{2}{*}{\makecell[c]{Resizer's\\ Input Size}} & \multirow{2}{*}{\makecell[c]{Resizer's\\ Output Size}} & \multirow{2}{*}{\makecell[c]{5-way \\1-shot}} & \multirow{2}{*}{\makecell[c]{5-way \\5-shot}}\\ \\
 \midrule 
    Baseline & original & 84 × 84 & 62.26 & 80.19 \\
    ASM & original & 84 × 84  & 64.41 & 82.34 \\ 
    MAR &  112 × 112 & 84 × 84  & 64.84  & 80.28 \\ 
    MAR & 126 × 126 & 84 × 84  & 64.45  & 80.23 \\ 
    MAR & 168 × 168 & 84 × 84  &  65.02 & 80.31 \\
    MAR+ASM &  112 × 112 & 84 × 84 & 64.10 & 82.00  \\ 
    MAR+ASM &  126 × 126 & 84 × 84  & 63.98 & 81.97  \\ 
    MAR+ASM &  168 × 168 & 84 × 84  & \textbf{65.19}  & \textbf{82.38} \\
\bottomrule   
\end{tabular} 
\caption{Results of different combinations of MAR and ASM on \emph{mini}-ImageNet.}  
\label{table:ablation} 
%3
\begin{tabular}{cccccc}   
\toprule
 \multirow{2}{*}{Model} & \multirow{2}{*}{\makecell[c]{Number of\\ MARBlocks}} & \multirow{2}{*}{\makecell[c]{5-way 1-shot\\(\%)}} & \multirow{2}{*}{\makecell[c]{5-way 5-shot\\(\%)}} \\\\
 \midrule 
    Ours & 2 & 63.73 ± 0.49 & 81.73 ± 0.34 \\
    Ours & 4 & \textbf{65.19 ± 0.52} & \textbf{82.38 ± 0.33} \\
\bottomrule   
\end{tabular}
\caption{Results of different number of MARBlocks on \emph{mini}-ImageNet. The input size of MAR is 168 × 168.} 
\label{table:MARBlock} 
\end{center}   
\end{table}
%BSNet Ablation
\begin{figure}[!ht]
\centering
\includegraphics[width=\linewidth]{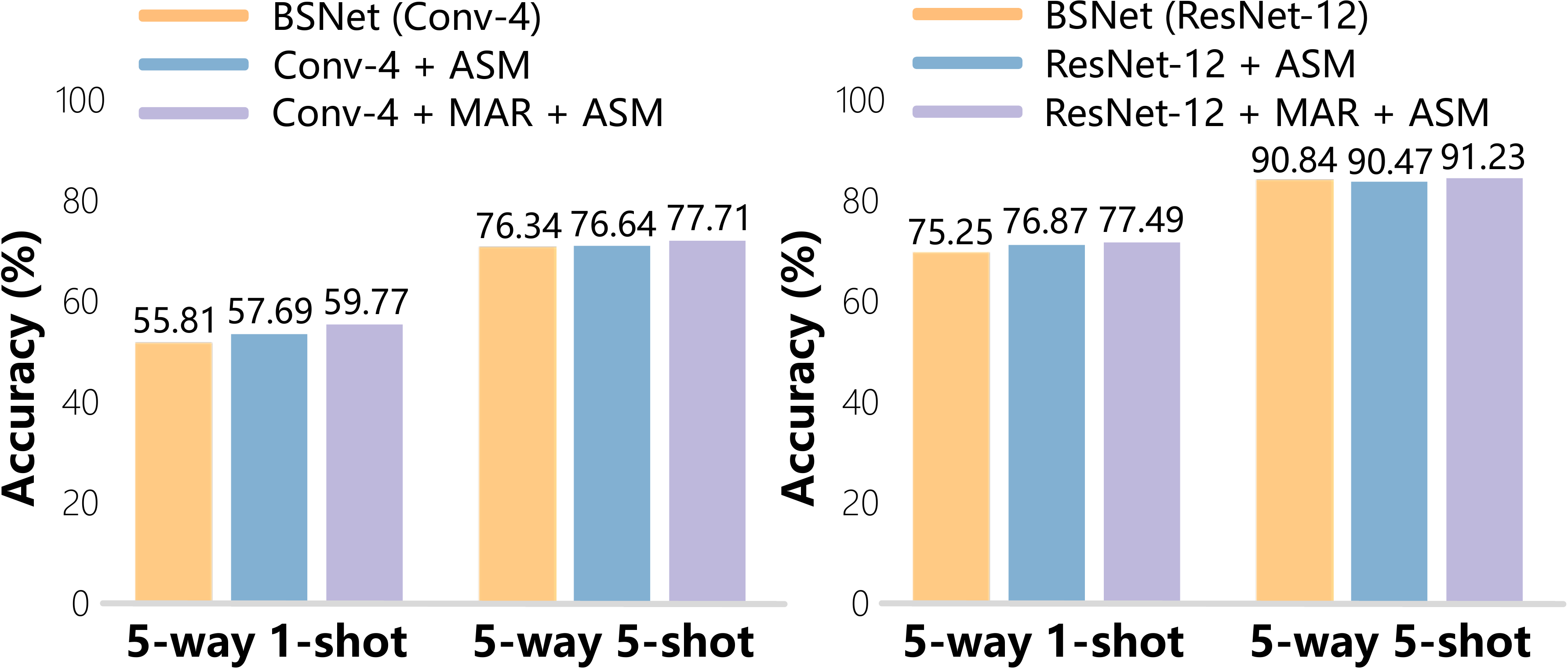}
\caption{Results of fine-grained FSL experiments compared with BSNet \cite{BSNet} on CUB.}
\label{fig:bsablation}
\end{figure}

\subsection{Experimental results}
For a fair comparison, a baseline of the proposed method has been built with ResNet-12 as backbone. Input size of MAR is set to 168 × 168 and weights of ASM, $\alpha$ and $\beta$, are initialized to 1.24 and 0.1. Results can be seen from Table \ref{table:sota}. On \emph{mini}-ImageNet and \emph{tiered}-ImageNet, this baseline improves 2.93\% and 2.19\%, 1.54\% and 3.1\% compared with \cite{ProtoNet}, and 0.28\% and 1.6\%, 3.24\% and 1.96\% compared with \cite{YinboChen2020MetaBaselineES}, at 5-way 1-shot and 5-shot settings, respectively. To further explore the plug-and-play property of the proposed model, we instantiate it in FEAT, Meta-Baseline and Meta DeepBDC. The results are shown in Table \ref{table:sota}. It improves FEAT by 0.74\% and 0.77\%, Meta-Baseline by 0.2\% and 3.04\%, Meta DeepBDC by 0.87\% and 1.03\% at 5-way 1-shot, 0.97\% and 0.94\%, 0.89\% and 2.13\%, 0.06\% and 0.04\% at 5-way 5-shot, on \emph{mini}-ImageNet and \emph{tiered}-ImageNet, respectively.
\subsection{Ablation analysis}
Extensive experiments have been done to demonstrate the effectiveness of the proposed method. Specifically, we build a baseline with ResNet-12. Three different configurations, MAR, ASM and MAR + ASM, are obtained by instantiating MAR and ASM. 112 x 112, 126 x 126, 168 x 168 are explored as different input sizes for MAR. The details are shown in Table \ref{table:ablation}. Compared with baseline, the addition of MAR and ASM improved the model at 5-way 1-shot and 5-way 5-shot by up to 2.93\% and 2.19\% on \emph{mini}-ImageNet. And MAR has a large improvement for 5-way 1-shot and ASM for 5-way 5-shot, reaching 2.76\% and 2.15\%, respectively.
To further explore the proposed method, we investigate the impact of different number of MARBlocks on the performance. Results can be seen in the Table \ref{table:MARBlock}. It shows that 4 MARBlocks are superior to 2. Moreover, we compare the proposed framework with BSNet \cite{BSNet}. Similar to ResNet-12, we also build a baseline with Conv-4. Based on them, we explore the performance of different configurations on CUB. Results are shown in Fig. \ref{fig:bsablation}. At 5-way 1-shot and 5-way 5-shot, our method improves up to 3.96\% and 1.37\% for Conv-4 and 2.24\% and 0.39\% for ResNet-12. And learned $\alpha$ and $\beta$ are 1.24 and 0.1, 1.91 and 0.1, respectively, which further explores BSNet. Since CUB is a fine-grained dataset, the results also demonstrate that a generalization of our approach to other domains. 

\section{Conclusion}
In this paper, we propose a plug-and-play framework for FSL. A learnable resizer (MAR) is adopted to enhance the embeddings, and adaptive metric (ASM) is used to obtain advanced discrimination capabilities. MAR alleviates the loss of detailed information in traditional preprocessing pipeline, and ASM converts simple metrics into an advanced and efficient one. In addition, we open a novel view to improve FSL from the perspective of input data and adaptive similarity measurement. Extensive experiments demonstrate that the proposed adaptive framework is a potential direction and could effectively boost existing methods.
% \vfill\pagebreak

\bibliographystyle{IEEEbib}
\bibliography{strings,refs}

\end{document}